\documentclass{article}

\PassOptionsToPackage{numbers, compress}{natbib}

\usepackage[preprint]{neurips_2024}

\usepackage{amsmath,amsfonts,bm}

\def\eqref#1{equation~\ref{#1}}

\def\1{\bm{1}}

\DeclareMathAlphabet{\mathsfit}{\encodingdefault}{\sfdefault}{m}{sl}
\SetMathAlphabet{\mathsfit}{bold}{\encodingdefault}{\sfdefault}{bx}{n}

\usepackage[utf8]{inputenc} %
\usepackage[T1]{fontenc}    %
\usepackage{hyperref}       %

\usepackage{url}            %
\usepackage{booktabs}       %
\usepackage{amsfonts}       %
\usepackage{nicefrac}       %
\usepackage{microtype}      %
\usepackage{xcolor}         %

\usepackage{xspace}

\usepackage{cleveref}

\usepackage{ifthen}

\newcommand{\todo}[1]{%
  \ifthenelse{\equal{#1}{}}%
    {\textcolor{purple}{\small \bf TODO}}%
    {\textcolor{purple}{\small \bf TODO: #1}}%
}

\newcommand{\sys}{{Scorch}\xspace}

\usepackage{amsmath}
\usepackage{amssymb}
\usepackage{mathtools}
\usepackage{amsthm}

\usepackage{graphicx}

\usepackage{subcaption}

\usepackage[noend]{algpseudocode}
\usepackage{algorithm}

\usepackage{semantic}

\usepackage{amssymb}
\usepackage{pifont}

\usepackage{caption}
\theoremstyle{plain}

\theoremstyle{definition}

\theoremstyle{remark}

\title{Scorch: A Library for Sparse Deep Learning}

\author{
  Bobby Yan\textsuperscript{\normalfont 1}, Alexander J. Root\textsuperscript{\normalfont 1}, Trevor Gale\textsuperscript{\normalfont 1},
  David Broman\textsuperscript{\normalfont 2,1},
  Fredrik Kjolstad\textsuperscript{\normalfont 1}
  \\
  \textsuperscript{1}Stanford University
  \ \
  \textsuperscript{2}KTH Royal Institute of Technology
  \\
  \textsuperscript{1}\texttt{\{bobbyyan,ajroot,tgale,broman,kjolstad\}@cs.stanford.edu}, \textsuperscript{2}\texttt{dbro@kth.se}
}

\begin{document}

\maketitle

\begin{abstract}
  The rapid growth in the size of deep learning models strains the capabilities of traditional dense computation paradigms.
  Leveraging sparse computation has become increasingly popular for training and deploying large-scale models, but existing deep learning frameworks lack extensive support for sparse operations.
  To bridge this gap, we introduce Scorch, a library that seamlessly integrates efficient sparse tensor computation into the PyTorch ecosystem, with an initial focus on inference workloads on CPUs.
  Scorch provides a flexible and intuitive interface for sparse tensors, supporting diverse sparse data structures.
  Scorch introduces a compiler stack that automates key optimizations, including automatic loop ordering, tiling, and format inference.
  Combined with a runtime that adapts its execution to both dense and sparse data, Scorch delivers substantial speedups over hand-written PyTorch Sparse (\texttt{torch.sparse}) operations without sacrificing usability.
  More importantly, Scorch enables efficient computation of complex sparse operations that lack hand-optimized PyTorch implementations. This flexibility is crucial for exploring novel sparse architectures.
  We demonstrate Scorch's ease of use and performance gains on diverse deep learning models across multiple domains. With only minimal code changes, Scorch achieves 1.05--5.78$\times$ speedups over PyTorch Sparse on end-to-end tasks.
  Scorch's seamless integration and performance gains make it a valuable addition to the PyTorch ecosystem. We believe Scorch will enable wider exploration of sparsity as a tool for scaling deep learning and inform the development of other sparse libraries.
\end{abstract}

\section{Introduction}

The rapid increase in model size and sophistication has driven the success of deep learning across domains like computer vision, natural language processing, and speech recognition.
This growth strains the capabilities of traditional dense computation paradigms, as model parameters often exceed available processing resources and memory.
Sparse computation techniques offer a promising solution by focusing compute and storage only on the most relevant elements~\cite{hoefler2021sparsity}.

Sparsity arises naturally in many contexts within deep learning and can generally be categorized into three groups: data sparsity, weight sparsity, and activation sparsity.
It arises in data, as many emerging application domains involve inherently sparse data representations.
For instance, real-world graphs often exhibit power-law distributions in node connectivity, yielding sparse adjacency matrices~\cite{gnn_survey_2021}. Similarly, the high-dimensional feature spaces used in recommender systems~\cite{DLRM19} and natural language processing~\cite{sparse_transformer_2019} are highly sparse.

Sparsity also arises within models, where weight and activation sparsity can provide significant efficiency benefits. Techniques like pruning~\cite{han2015deep, blalock2020state, luo2017thinet} have shown that weights can be sparsified with little impact on accuracy. Sparse architectures, such as mixture-of-experts (MoEs)~\cite{shazeer_iclr_2017, lepikhin2020gshard, fedus2022switch} and sparse transformers~\cite{sparse_transformer_2019, beltagy2020longformer, kitaev2020reformer}, sparsify activations using sparsely-gated conditional computation, improving model capability without proportional increases in the memory or computational cost.

However, it is challenging for researchers and end users to take full advantage of sparsity in deep learning due to limitations in existing deep learning frameworks.
Mainstream frameworks like PyTorch~\cite{pytorch_2019}, JAX~\cite{jax2018github}, and TensorFlow~\cite{tensorflow_2016} provide sparse support for only a handful of operations on a few selected sparse data structures (tensor formats).
As a result, leveraging sparsity in these frameworks often demands significant engineering effort to develop custom implementations for new operations.
These frameworks lack the flexibility for researchers to explore the diverse range of potentially beneficial sparse data structures and computations in modern deep learning architectures.

Indeed, nearly every domain where sparsity arises has one-off libraries with hand-written and hand-optimized kernels, such as PyTorch Geometric~\cite{pyg_2019} for graph learning, TensorLy~\cite{tensorly} for sparse tensor factorization,
and MegaBlocks~\cite{megablocks2023} for MoEs.
Although these libraries effectively target their specific applications, they do not generalize to other domains, resulting in a fragmented software ecosystem for sparse deep learning.
The different abstractions create barriers to adoption, as they require users to learn new APIs and paradigms for each library.
This also results in duplicated developer effort, as each domain-specific library implements its own sparse kernels and optimizations.
Therefore, there is an opportunity for a unified framework that efficiently handles sparse operations and representations across various domains in deep learning.

To tackle these challenges, we introduce Scorch\footnote{Scorch is available under the MIT license at \url{https://anonymous.4open.science/r/scorch}.}, the first library to integrate comprehensive and efficient sparse tensor computation capabilities into PyTorch.
Scorch adds sparsity support to existing PyTorch operations, allowing researchers and practitioners to introduce sparsity into their models by simply declaring one or more tensors to be sparse. For instance, weight matrices can be made sparse, sparse gating can be used in architectures like MoEs for conditional computation, and sparse adjacency matrices can be efficiently processed in graph neural networks (GNNs). By preserving PyTorch's interface, Scorch enables the natural and seamless expression of sparse models without requiring a separate ecosystem of sparse operations.
Our initial work focuses on accelerating compute operations on CPUs, which enables sparse inference workloads and lays a foundation for general sparse computing in PyTorch. We leave GPU acceleration and auto-differentiation as future work.

The key insight behind Scorch is that it is possible to provide comprehensive sparsity support with minimal API changes to existing tensor frameworks. Sparsity is just a property of the values of a tensor and should not require a separate programming model. opt  that everything doable with dense tensors should also be possible with sparse tensors. Scorch is designed to enable this functionality with a single line of code: {\small\texttt{import scorch as torch}}.
Scorch enables general sparsity support by integrating state-of-the-art sparse code generation infrastructure~\cite{taco_2017,format_2018,senanayake2020scheduling} into a modern ML framework.
However, prior work on sparse compilation lacks several technical components necessary to achieve such minimal API changes without sacrificing performance.
We identify several key missing components and make the following technical contributions:
\begin{itemize}
    \item A fast auto-scheduler for optimizing loop ordering, fusion, and the insertion of temporary data structures in the generated sparse kernels.
    \item A tiling algorithm that optimizes sparse compute kernels to improve cache locality.
    \item A format inference algorithm that determines the output format for each operation based on the nature of the operation and the input formats.
\end{itemize}

We evaluate Scorch on models across a range of domains, including graph neural networks, sparse autoencoders, and sparse transformers. We demonstrate speedups of 1.05--5.78$\times$ over PyTorch Sparse.
This paper illustrates how to design a high-performance end-to-end sparse ML library.
By enabling the natural expression and efficient execution of sparse models, Scorch can facilitate greater research exploration and adoption of sparsity.

\section{Related Work}
\label{sec:related_work}

\textbf{Deep learning frameworks.}
Several existing frameworks aim to support deep learning models for training and inference, but native sparse support in mainstream frameworks like PyTorch~\cite{pytorch_2019}, JAX~\cite{jax2018github}, and TensorFlow~\cite{tensorflow_2016} is limited in generality, efficiency, and usability.
Their sparsity support focuses on select formats like COO (coordinate list) and CSR (compressed sparse row), and a limited number of sparse operations like SpMM and SpMV.
While libraries like MKL-Sparse and cuSPARSE provide low-level sparse primitives, they do not support high-order tensor operations, most formats across all operations, or fusion.
Consequently, libraries that build support for general sparse tensor operation on top of them suffer from excessive data movement, which leads to poor performance.

Domain-specific libraries like PyTorch Geometric (PyG)~\cite{pyg_2019} and Deep Graph Library (DGL)~\cite{dgl_2019} have been developed to target sparse computation in specific applications. While they provide powerful capabilities within their target domains, they lack the generality to support other applications and sparse data structures. Scorch introduces a unified sparse tensor abstraction and compiler infrastructure to enable efficient and composable sparse deep learning across multiple domains.

\textbf{Tensor algebra and machine learning compilation.}
Many compiler frameworks have been proposed for optimizing tensor operations and neural networks, such as Halide~\cite{halide_2013}, TVM~\cite{tvm_2018}, XLA~\cite{xla}, Glow~\cite{glow_2018}, and Tensor Comprehensions~\cite{tensor_comprehensions_2018}. These compilers introduce new intermediate representations (IRs) and tensor algebra compilation techniques.
However, they primarily focus on dense tensor computations and lack first-class support for sparse code generation.

Recent work on sparse tensor algebra compilers, such as TACO~\cite{taco_2017}, Tiramisu~\cite{tiramisu_2019}, MLIR Sparse~\cite{mlir_sparse_2022}, and SparseTIR~\cite{sparseTIR_2023}, has made significant progress in code generation for sparse tensor algebra expressions.
However, these compilers are not integrated with deep learning frameworks, making it challenging to use them for sparse deep learning workloads.
They also lack efficient and general automatic optimization algorithms necessary for high-performance sparse computation.
\sys incorporates ideas from the sparse code generation literature into the PyTorch ecosystem and adds automatic optimizations machinery to provide a seamless, end-to-end approach to productive and performant sparse deep learning.

\section{Design}

Scorch's design is guided by three principles that aim to balance performance, usability, and generality in the context of sparse deep learning:

\textbf{Seamless integration with PyTorch.}
To facilitate adoption and usability, a sparse deep learning library must integrate seamlessly with popular deep learning frameworks.
Scorch is designed as a natural extension of PyTorch, with a familiar and intuitive interface for defining and manipulating sparse tensors.
Users can leverage their existing PyTorch knowledge and codebases and easily switch between dense and sparse computation. This enables smooth adoption of Scorch and allows researchers to focus on developing new models rather than learning new frameworks.

\textbf{Unified abstractions for dense and sparse computation.}
Deep learning computations often involve a mix of dense and sparse tensors, but existing frameworks provide limited support for sparse tensors in their operators. For instance, PyTorch's \texttt{einsum} and \texttt{reshape} operations do not support sparse tensors. To address this limitation, a sparse library should offer comprehensive operators that can freely operate on any mix of dense and sparse tensors, allowing users to express complex computations in natural tensor notation. This includes element-wise and non-linear operations, linear algebra, einsum operations, and tensor manipulations like {\texttt{reshape}} and \texttt{transpose}. Scorch accomplishes this by extending popular PyTorch operations to support sparse tensors. By providing unified operator support, Scorch lets users leverage the full power of PyTorch's API while taking advantage of efficient sparse computation under the hood.

\textbf{Robust automatic performance optimization.}
Achieving high performance on sparse workloads requires careful co-optimization of storage, loop orderings, use of intermediates, and tiling schemes. These choices collectively determine both the asymptotic complexity of a sparse kernel and the constant factors contributing to the runtime. The compiler stack of Scorch is capable of making reasonable decisions across this scheduling space using a set of carefully designed heuristics. Scorch's optimization algorithms are designed around the following hierarchy of goals: First, format inference and loop ordering decisions should match those implemented in the state-of-the-art hand-written kernels. Second, loop ordering and storage decisions should ensure good worst-case asymptotic complexity to avoid unpredictable performance cliffs. Third, tiling should be applied to dense loops within sparse kernels to optimize cache usage. This heuristic hierarchy enables Scorch to match the performance of existing hand-written kernels and achieve predictable performance on unknown kernels, without burdening the user with manual performance tuning.

\subsection{Programming Model}

\begin{figure}
  \centering
  \begin{minipage}[b]{0.61\linewidth}
    \includegraphics[width=\linewidth]{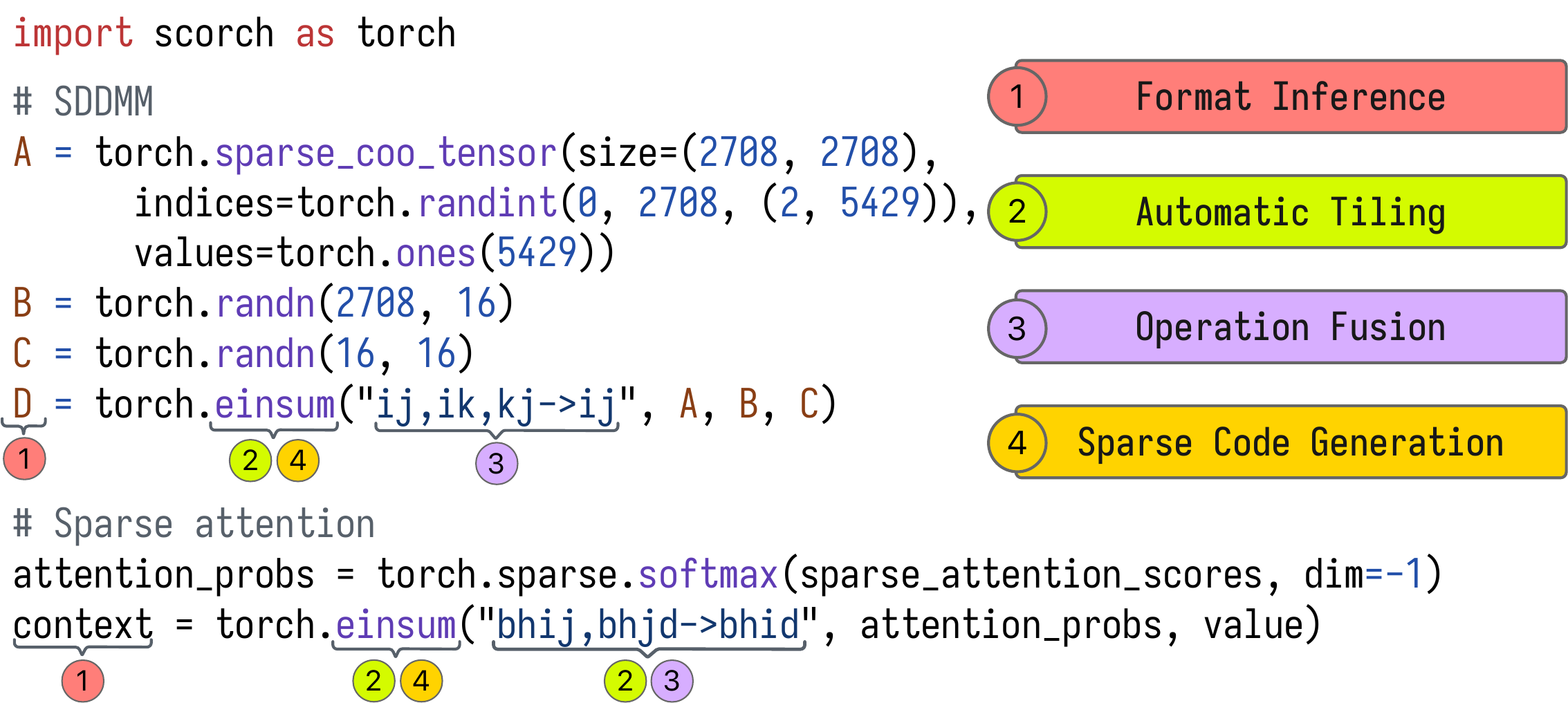}
    \caption{Scorch SDDMM and sparse attention examples}
    \label{fig:example}
  \end{minipage}
  \hfill
  \begin{minipage}[b]{0.34\linewidth}
    \includegraphics[width=\linewidth]{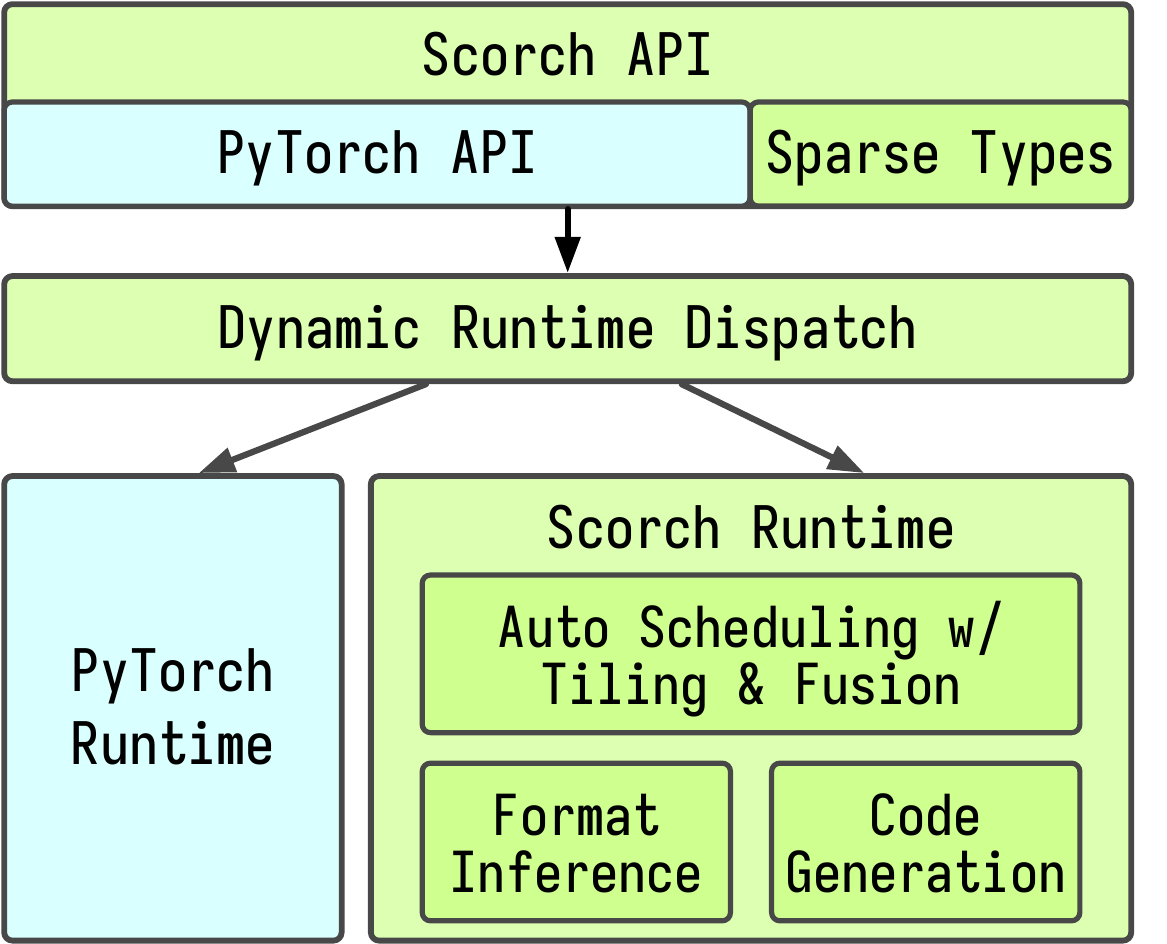}
    \caption{\sys architecture}
    \label{fig:overview}
  \end{minipage}
\end{figure}

Scorch introduces a unified programming model for sparse and dense deep learning. Users opt into sparse execution via an import statement that make standard PyTorch operations compatible with sparse tensors.
\Cref{fig:example} illustrates the simplicity and flexibility of Scorch's programming model and how it improves productivity without sacrificing performance.
The first example shows a Sampled Dense-Dense Matrix Multiplication (SDDMM) operation, commonly used in GNNs and recommender systems.
It takes a sparse matrix $A$ and two dense matrices $B$ and $C$ as inputs and computes the sparse matrix $D$ as the element-wise product of $A$ and the matrix multiplication of $B$ and $C$.
PyTorch does not support fused SDDMM for sparse COO inputs due to the lack of a hand-written kernel.
Scorch enables users to express this computation concisely using \texttt{einsum}. By leveraging its compiler, Scorch automatically generates an efficient fused kernel for SDDMM.
The generated kernel co-iterates over the sparse and dense tensors, avoiding the asymptotically more expensive dense matrix multiplication in the worst case.
As seen in this example, fusion is crucial for performance, as it can reduce the overall computational complexity.
By supporting mixed sparse and dense inputs operators like \texttt{einsum}, Scorch unlocks a wide range of sparse operations that would be difficult or impossible to express using PyTorch's fixed set of hand-written sparse kernels.

\subsection{System Overview}

\Cref{fig:overview} shows an overview of the architecture of Scorch. Its compiler lowers high-level sparse tensor operations into optimized kernels for efficient execution. Compilation begins by transforming Python code to a domain-specific intermediate representation (IR) that exposes optimization opportunities. The compiler then applies the following optimization passes to the IR:

\begin{itemize}
  \item \textbf{Automatic scheduling}: Applies loop nest optimizations, tiling, and parallelization techniques specifically tuned for sparse tensor computations.
  \item \textbf{Format inference}: Automatically determines the most suitable output storage format for each operation based on its inputs and the operation.
  \item \textbf{Code generation}: Translates the optimized IR into efficient low-level kernels specialized for the specific storage formats of the input and output tensors.
\end{itemize}

A lightweight dispatcher orchestrates the execution of kernels. For common operations like SpMM, the dispatcher retrieves pre-compiled kernels from cache to save compilation time. For kernels that are not in the cache, the dispatcher invokes the above compiler at runtime to generate C++ code, which is then compiled and dynamically linked into the library.

\section{Optimizations}

Scorch introduces several optimizations to ensure the generated code has good and predictable performance.
Efficient execution of sparse tensor operations requires careful consideration of the loop order, tiling, and the use of temporary tensor data structures. Suboptimal choices can lead to poor cache utilization, inefficient memory access patterns, and unnecessary data movement. In fact, the wrong loop order in a sparse loop nest can even change its asymptotic complexity~\cite{ahrens_2022}.  To address this challenge, \sys introduces new algorithms for auto-scheduling sparse kernels that determines the loop order, workspace insertion, and tiling strategy.

\subsection{Automatic Loop Ordering and Workspace Insertion}
\label{sec:loop-ordering}

Unlike dense computations, where different loop orders have the same asymptotic complexity, the choice of loop order can impact the worst-case asymptotic complexity and thus the observed performance of sparse computations~\cite{ahrens_2022}. Our loop ordering algorithm is designed to provide predictable performance to users by avoiding loop orders with poor asymptotic complexity.
However, optimizing solely for the worst-case complexity may lead to bad average-case performance. We provide a fast heuristic auto-scheduler that achieves good average-case performance while avoiding the most asymptotically inefficient loop orderings.

\begin{figure}
\centering
\begin{minipage}[b]{0.34\linewidth}
\includegraphics[width=\linewidth]{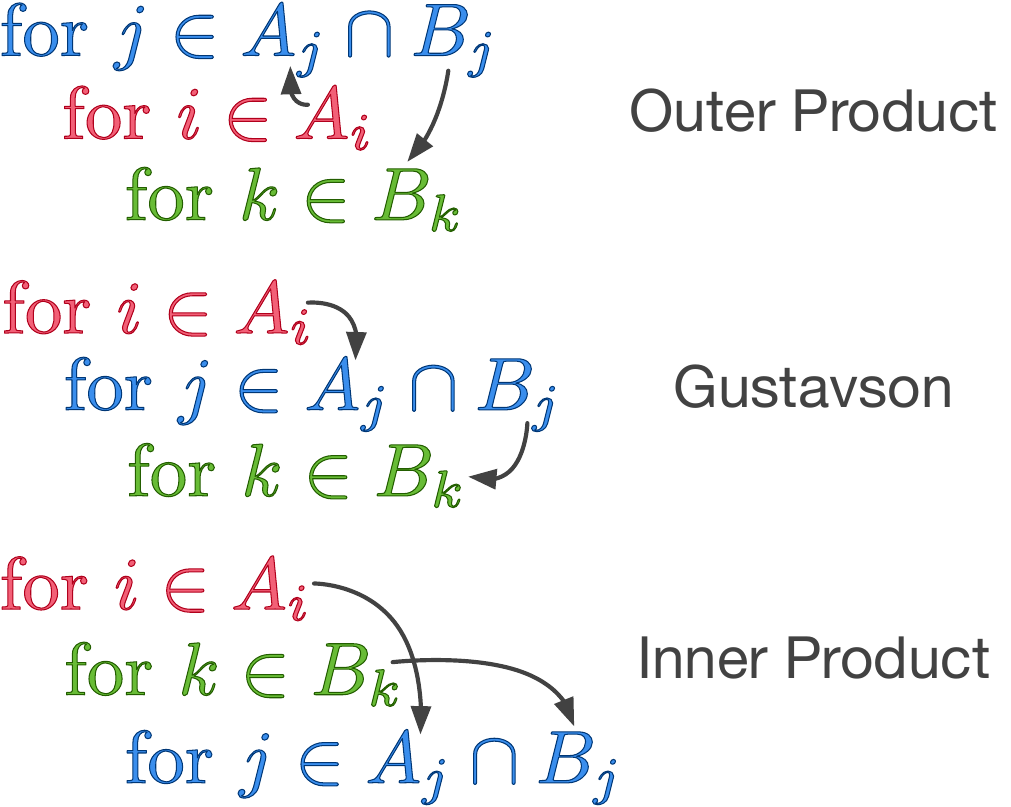}
\caption{SpGEMM loop orders}
\label{fig:loop-ordering}
\end{minipage}
\hfill
\begin{minipage}[b]{0.45\linewidth}
\includegraphics[width=\linewidth]{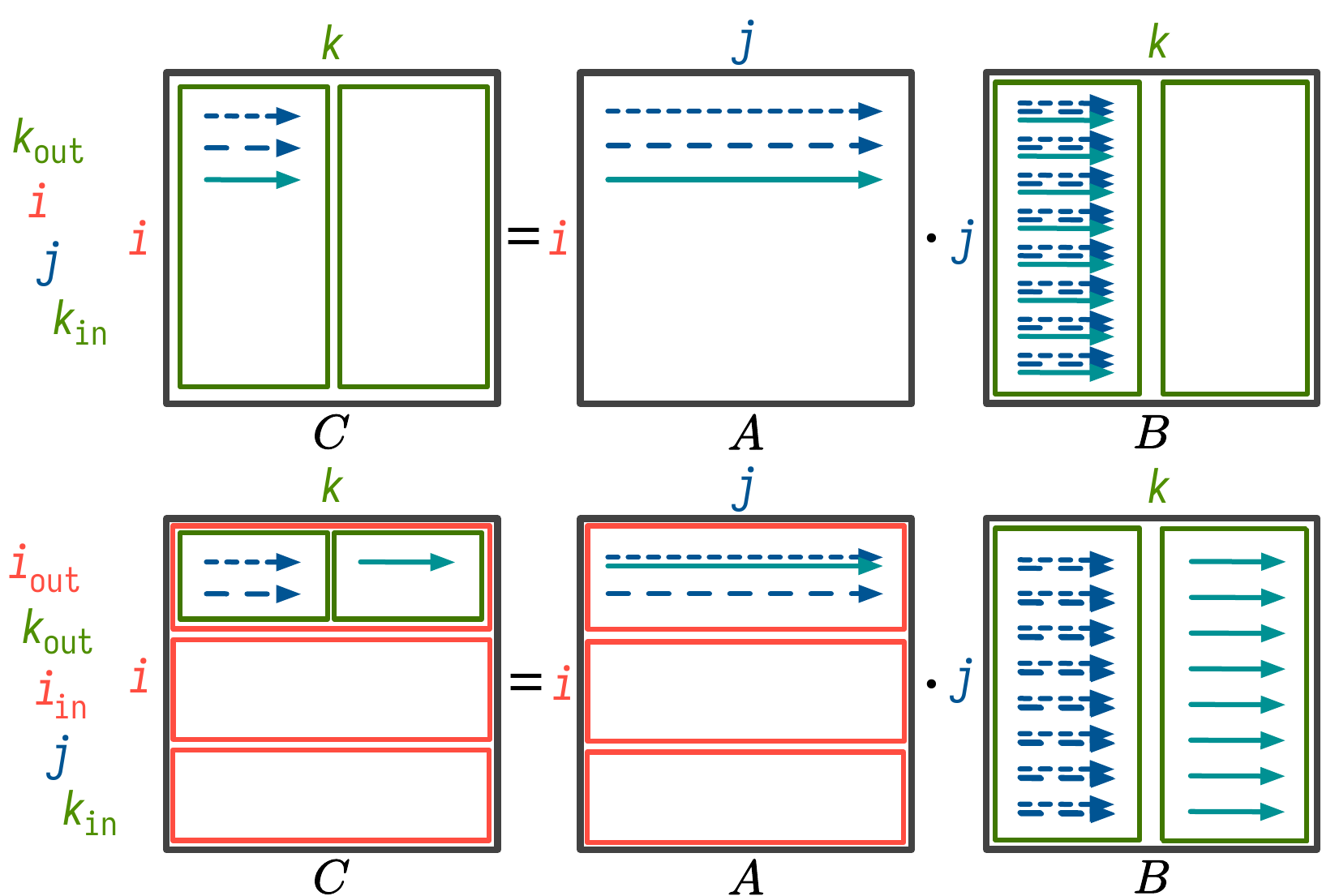}
\caption{Tiling SpMM (\(C_{ik} = \sum_j A_{ij} B_{jk}\))}
\label{fig:spmm-tiling}
\end{minipage}
\end{figure}

Consider the sparse matrix-matrix multiplication (SpGEMM), $C_{ik} = \sum_j A_{ij} B_{jk}$, where the inputs $A$, $B$ and the output $C$ are stored in CSR format. \Cref{fig:loop-ordering} illustrates three possible loop orderings for SpGEMM: outer product, Gustavson, and inner product. The loops iterate over the coordinates of different dimensions of the matrices, and the arrows denote the data structures for each matrix. For example, the $A$ matrix in Gustavson stores $i$ coordinates (denoted $A_i$) followed by $j$ coordinates (denoted $A_j$)\footnote{In \Cref{fig:loop-ordering}, $T_{i}$ denotes the set of coordinates in the dimension of tensor $T$ indexed by index variable $i$.}. The inner product algorithm has worse asymptotic complexity than the other two algorithms because the data structures of neither $A$ nor $B$ connects the coordinates in the two outer loops. They thus have to be iterated densely, meaning every component of the result has to be visited, even those that result in zeros. The other algorithms do not have this problem and thus perform much better. Moreover, algorithms that perform intersections between sparse matrices in higher loops (early filtering) perform better than those that perform intersections in inner loops. Scorch's loop ordering algorithm avoids bad asymptotic behavior, even at the expense of transposes to allow matrices to be iterated in different orders.

Our algorithm (\Cref{alg:looporder}) first collects and sorts index variables in the tensor expression by descending sparsity level. The sparsity level of an index variable is approximated by the presence of intersections (filters) between levels. In this case, $j \in A_j \cap B_j$ is the only sparse filter as it involves an intersection between the sparse level $A_j$ and the dense level $B_j$. The tie between $i$ and $k$ are broken by choosing the order that requires fewer transposes. This gives an initial loop order $\mathcal{L} = [j, i, k]$.

The algorithm then takes a greedy approach and iteratively evaluates the cost of pushing a sparse filter down the loop nest. In the SpGEMM example, the algorithm considers the net cost of moving $j$ to $\textit{pos}=1$: the filter no longer applies to the rows of $A$ in the $i$ loop, but the 2D workspace can be replaced by a 1D workspace since we no longer scatter into the $i$ dimension, and $A$ no longer needs to be transposed to iterate in $j, i$ order. As the net cost is negative, the loop order is updated to $[i, j, k]$. In the next iteration, we consider the net cost of moving $j$ to $\textit{pos}=2$, i.e., the inner product order $[i, k, j]$: while the 1D workspace can be eliminated since we no longer scatter into the result, the filter no longer applies to any of the tensors, and $B$ needs to be transposed to be iterated in the $k, j$ order.
Moreover, as there are no input data structures that allow us to iterative sparsely from $i$ to $k$, it incurs a large penalty in the cost function. This makes net cost positive, so the loop order remains unchanged.

\begin{algorithm}
  \caption{Loop Ordering and Workspace Insertion}
  \label{alg:looporder}
  \begin{algorithmic}[1]
    \State \textbf{Input:} Tensor expression ${E}$, input tensors $\mathcal{T}$, output tensor $T_{out}$.
    \State \textbf{Output:} Loop order $\mathcal{L}$ for efficient computation.
    \State $\mathcal{I} \gets \textsc{GetIndexVariables}(E)$ \Comment{Set of all index variables in the expression}
    \State $\mathcal{S} \gets \textsc{SortBySparsity}(\mathcal{I}, E)$ \Comment{Sort index variables by descending sparsity}
    \State $\mathcal{L} \gets \textsc{InitLoopOrder}(\mathcal{S})$

    \ForAll{$i \in \mathcal{S}$} \Comment{Loop ordering}
      \State $\textit{currentPos} \gets \textsc{GetPosition}(\mathcal{L}, i)$ \Comment{Current position of index $i$ in $\mathcal{L}$}
      \For{\textit{pos} $\gets \textit{currentPos}$ \textbf{to} $|\mathcal{L}| + 1$}

        \If{\textsc{Cost}($\mathcal{L}, i, \textit{pos}) < 0$}
        \Comment{Net cost of pushing $i$ to position \textit{pos}}
          \State $\mathcal{L} \gets \textsc{MoveToPosition}(\mathcal{L}, i, \textit{pos})$
        \EndIf
      \EndFor
    \EndFor

    \State $G \gets \textsc{InitGraph(}\mathcal{I}\textsc{)}$ \Comment{Initialize a directed graph with nodes $\mathcal{I}$}
    \ForAll{$T \in \mathcal{T} \cup \{T_{out}\}$}
      \State $\pi \gets \textsc{GetModeOrdering(}T\textsc{)}$ \Comment{Get the mode ordering of tensor $T$}
      \For{$k \gets 1$ \textbf{to} $|\mathcal{I}(T)| - 1$}
        \State $\textsc{AddEdgeToGraph(}G, \pi[k], \pi[k+1]\textsc{)}$ \Comment{Add edge to capture ordering constraint}
      \EndFor
    \EndFor

    \While{$\textsc{ContainsCycles(}G\textsc{)}$}
      \State $\textsc{RemoveCheapestEdge}(G)$ \Comment{Remove the cheapest edge from the graph}
      \State $\mathcal{L} \gets \textsc{UpdateLoopOrder(}\mathcal{L}, e\textsc{)}$ \Comment{Update the loop order by transposing the tensor}
    \EndWhile

    \If{$\textsc{HasSparseLevels(}T_{out}\textsc{)}$} \Comment{Workspace insertion}
      \State $\mathcal{I}_{red} \gets \textsc{GetReductionVariables(}\mathcal{T}, T_{out}\textsc{)}$ \Comment{Get reduction variables}
      \If{$\textsc{ShouldInsertWorkspace(}\mathcal{I}_{red}, \mathcal{I}(T_{out}), \mathcal{L}\textsc{)}$}
        \State $W \gets \textsc{InsertWorkspace(}i\textsc{)}$ \Comment{Insert a workspace over loop $i$}
        \State $(\mathcal{L}_p, \mathcal{L}_c) \gets \textsc{SplitLoopOrder(}\mathcal{L}, i\textsc{)}$ \Comment{Split the loop order}
        \State $\mathcal{L}_p \gets \textsc{UpdateProducerLoop(}\mathcal{L}_p, W, E\textsc{)}$ \Comment{Update the producer loop}
        \State $\mathcal{L}_c \gets \textsc{UpdateConsumerLoop(}\mathcal{L}_c, T_{out}, W\textsc{)}$ \Comment{Update the consumer loop}
      \EndIf
    \EndIf

    \State \Return $\mathcal{L}$
  \end{algorithmic}
\end{algorithm}

\subsection{Automatic Tiling}
\label{sec:tiling}

Tiling is a crucial optimization that improves cache utilization and reduces memory traffic by partitioning the iteration space into smaller blocks (tiles) that fit in the cache. Selecting which loops to tile is challenging, as suboptimal choices may hurt performance. \sys introduces a novel sparse tiling algorithm that analyzes the tensor expression to determine which loops to tile based on several key observations:

First, opportunities for tiling can be inferred from the index variables present in each tensor access. In the SpMM example $C_{ik} = \sum_j A_{ij} B_{jk}$, there are three tensor accesses: $C[i, k]$, $A[i, j]$, and $B[j, k]$. If a tensor access is missing an index variable present in the full expression, it means that tensor access is reused across the loop corresponding to the missing index. For instance, $C[i, k]$ is missing $j$, indicating that it is reused across the $j$ loop. Thus, tiling the $i$ and $k$ loops would improve cache locality for $C$ if everything were dense. Second, tiling sparse dimensions is often counterproductive, as it requires performing expensive searches in the sparse data structures. Although it may be beneficial in highly tuned systems, Scorch avoids it in order to provide robust and predictable performance. In SpMM, for a CSR input $A$, the $j$ loop should not be tiled since $A_j$ is a sparse dimension. Third, not all dense dimensions should be tiled, as illustrated in \Cref{fig:spmm-tiling}. While tiling $i$ in addition to $k$ would improve reuse of the rows of $A$, it would also require a larger working set for $B$ since the tiles of $k$ would be exhausted more frequently. In the dense case, the $j$ loop can be tiled to offset this, but since sparse dimensions are not tiled, tiling $i$ should be avoided. In general, tiling loops whose index variables are parent index variables of a sparse dimension should be avoided.

These insights form the basis of \sys's tiling algorithm, which can be summarized as follows: First, for each tensor access, collect its index variables if the set of index variables in the tensor access is a strict subset of the set of all index variables in the expression, as they indicate data reuse. Then, remove any index variables that correspond to a sparse dimension of any tensor. Finally, loops whose index variables are parents of a sparse index variable should not be tiled.  The full algorithm is provided in \Cref{app:tiling-algo}.

\subsection{Format Inference}

Tensor format inference is another key optimization in \sys that automatically determines the storage format of the output tensor based on the storage formats of the input tensors and the semantics of the tensor expression. By inferring storage formats amenable to good performance, \sys can generate efficient sparse kernels without requiring the user to manually specify the output format.

\sys's tensor format inference algorithm operates on a per-dimension basis, considering the interaction between the storage formats of each dimension in the input tensors. The intuition behind the algorithm is that multiplying a sparse level with any other level (sparse or dense) results in a sparse level in the output, while adding a dense level to any other level yields a dense level in the output. \Cref{app:format-inference} contains details of the algorithm and its theoretical properties.

\section{Evaluation}

In this section, we demonstrate that Scorch adds general sparse support to PyTorch and achieves performance comparable to or better than hand-tuned sparse kernels on end-to-end models across several domains: graph neural networks, sparse autoencoders, and sparse transformers.
We perform the benchmarks on an Apple M1 Ultra CPU (3.2 GHz, 20 cores) with 64 GB of memory.

\subsection{Graph Neural Networks}

We evaluate Scorch on the task of node classification using Graph Convolutional Networks (GCNs)~\cite{kipf2016semi}. GCNs are a popular class of graph neural networks that learn node representations by iteratively aggregating features from neighboring nodes. The core operation in a GCN layer is:
\begin{equation}
    \mathbf{H}^{(l+1)} = \sigma(\hat{\mathbf{A}} \mathbf{H}^{(l)} \mathbf{W}^{(l)})
\end{equation}
where $\mathbf{H}^{(l)}$ are the node features at layer $l$, $\hat{\mathbf{A}}$ is the normalized sparse adjacency matrix, $\mathbf{W}^{(l)}$ are the learnable weights, and $\sigma$ is a non-linear activation function.

We compare the inference performance of Scorch against implementations in PyTorch, PyG, and DGL on four benchmark datasets: Cora~\cite{cora_2000}, Citeseer~\cite{citeseer_1998}, PubMed~\cite{pubmed_2008}, and OGBN-arXiv~\cite{ogbn_2020}.

\begin{figure}
  \centering
  \includegraphics[width=\linewidth]{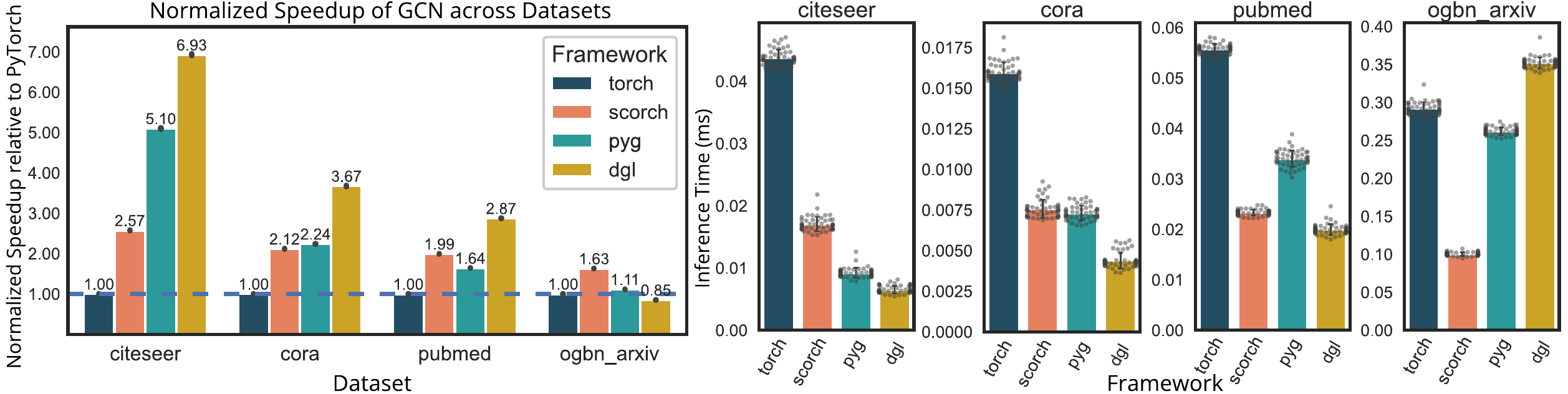}
  \caption{GCN Performance}
  \label{fig:gcn-perf}
\end{figure}

\Cref{fig:gcn-perf} shows the inference times of the GCN model and the normalized speedups relative to PyTorch on the four datasets. Scorch achieves an average speedup of 2.08$\times$.
The speedup achieved by Scorch can be attributed to several key optimizations in its SpMM kernel. Scorch applies tiling to improve cache locality and reduce memory traffic. It also uses pragma unroll directives to encourage loop unrolling, which can help reduce loop overhead and improve instruction-level parallelism.

Notably, the performance of PyG and DGL relative to PyTorch varies with the problem size. For smaller datasets like Citeseer and Cora, PyG and DGL achieves significant speedups over PyTorch. However, as the problem size grows, their performance advantage diminishes, for larger datasets like PubMed and OGBN-arXiv. This can be explained by the different algorithms used by these libraries for neighborhood aggregation. Instead of using SpMM directly, the gather-scatter algorithm indexes the source node features with the edge indices and scatters results into the destination nodes. While the gather-scatter algorithm is faster than SpMM for smaller graphs, it does not scale well to larger problems, as evident in the decreasing normalized speedups of PyG and DGL relative to PyTorch.

\subsection{Sparse Autoencoders}
We evaluate the performance of sparse autoencoders implemented using Scorch and compare it against the PyTorch implementation.
Autoencoders are unsupervised learning models that learn efficient data representations by encoding the input into a lower-dimensional latent space and then reconstructing the original input from the encoding. Sparse autoencoders introduce sparsity in the latent space, which can improve the interpretability and generalization of the learned representations~\cite{ng2011sparse}. The sparsity is typically achieved through regularization techniques like L1 penalty or KL-divergence loss. We train and test the autoencoders on four benchmark datasets: MNIST~\cite{lecun1998mnist}, CIFAR-10~\cite{krizhevsky2009learning}, CIFAR-100~\cite{krizhevsky2009learning}, and CelebA~\cite{liu2015deep}.

The sparse autoencoder architecture consists of an encoder with a sparse linear layer followed by ReLU activation, and a decoder with a dense linear layer followed by sigmoid activation. The models are trained using the mean squared error (MSE) loss and the Adam optimizer with a learning rate of 0.01. We train the models for 10 epochs with a batch size of 64.

\begin{figure}
  \centering
  \includegraphics[width=\linewidth]{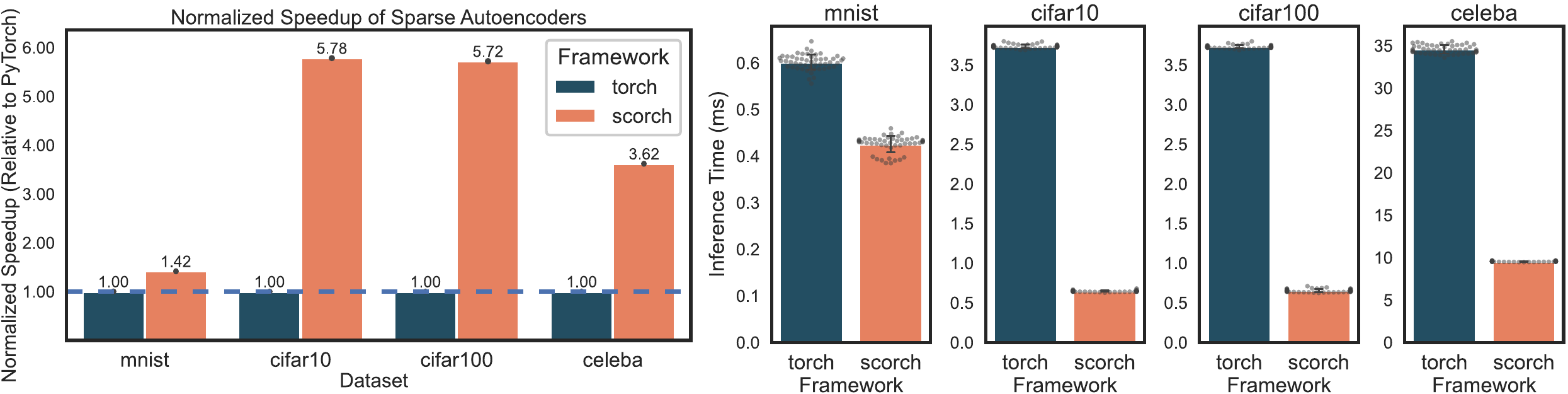}
  \caption{Sparse Autoencoder Performance}
  \label{fig:sparse_autoencoder_results}
\end{figure}

\Cref{fig:sparse_autoencoder_results} shows the speedup achieved by Scorch over PyTorch for sparse autoencoder inference on the different datasets. Scorch achieves speedups of 1.42$\times$ to 5.78$\times$ over PyTorch, with the speedup being more pronounced on larger datasets like CIFAR-100 and CelebA. This is because the sparse linear layer in the encoder dominates the computation time, and Scorch's optimized sparse kernels provide acceleration over PyTorch's sparse linear algebra routines.

\subsection{Sparse Transformers}

\begin{figure}
  \centering
  \includegraphics[width=\linewidth]{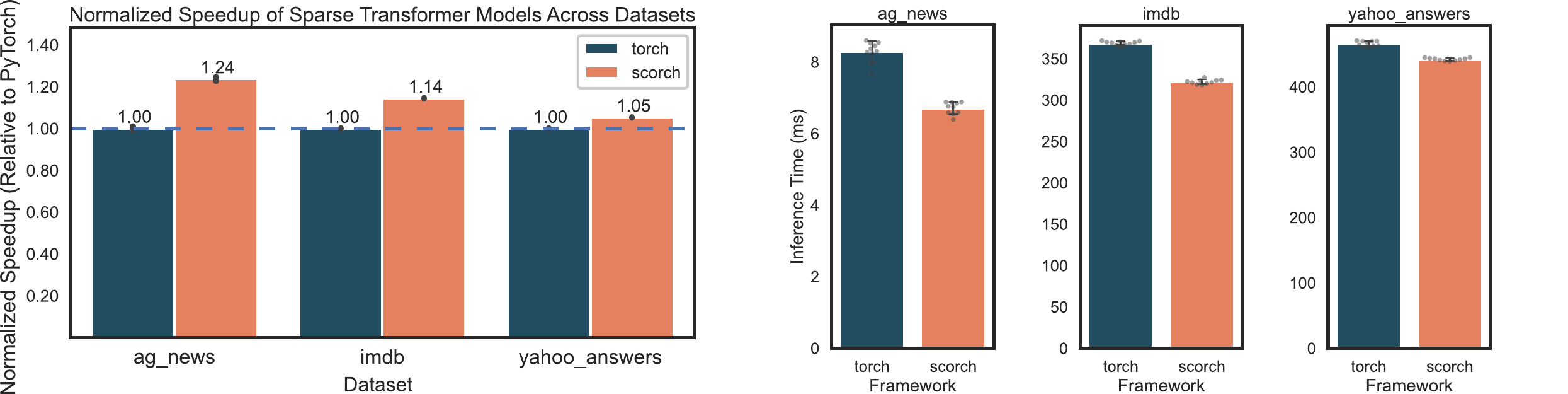}
  \caption{Sparse Transformer Performance}
  \label{fig:bigbird}
\end{figure}

We evaluate Scorch on the task of text classification using sparse transformer models, specifically the BigBird architecture~\cite{zaheer2020bigbird}. BigBird leverages sparse attention mechanisms, including global, window, and random attention, to scale to long sequences with linear memory complexity. We benchmark the inference performance of BigBird implemented using Scorch versus the standard dense PyTorch implementation on three datasets: IMDB~\cite{maas2011learning}, AG News~\cite{zhang2015character}, and Yahoo Answers~\cite{zhang2015character}.

\Cref{fig:bigbird} shows the inference times of BigBird. Scorch achieves an average speedup of 1.24$\times$, 1.14$\times$, and 1.05$\times$ in wall clock time for the three datasets respectively compared to PyTorch. Scorch's ability to generate efficient kernels for sparse attention operations allows it to accelerate inference on real-world text classification tasks with varying sequence lengths and sparsity patterns.

\subsection{Standard Kernels}

We evaluate the performance of \sys on several standard sparse kernels: Sparse Matrix-Vector Multiplication (SpMV), Sparse Matrix-Matrix Multiplication (SpMM), Sparse Matrix-Sparse Matrix Multiplication (SpMSpM), and Sampled Dense-Dense Matrix Multiplication (SDDMM). These operations are fundamental building blocks in many sparse deep learning models. We benchmark on matrices from the SuiteSparse Matrix Collection~\cite{suitesparse2011}.

\begin{figure}
  \begin{center}
    \includegraphics[width=\linewidth]{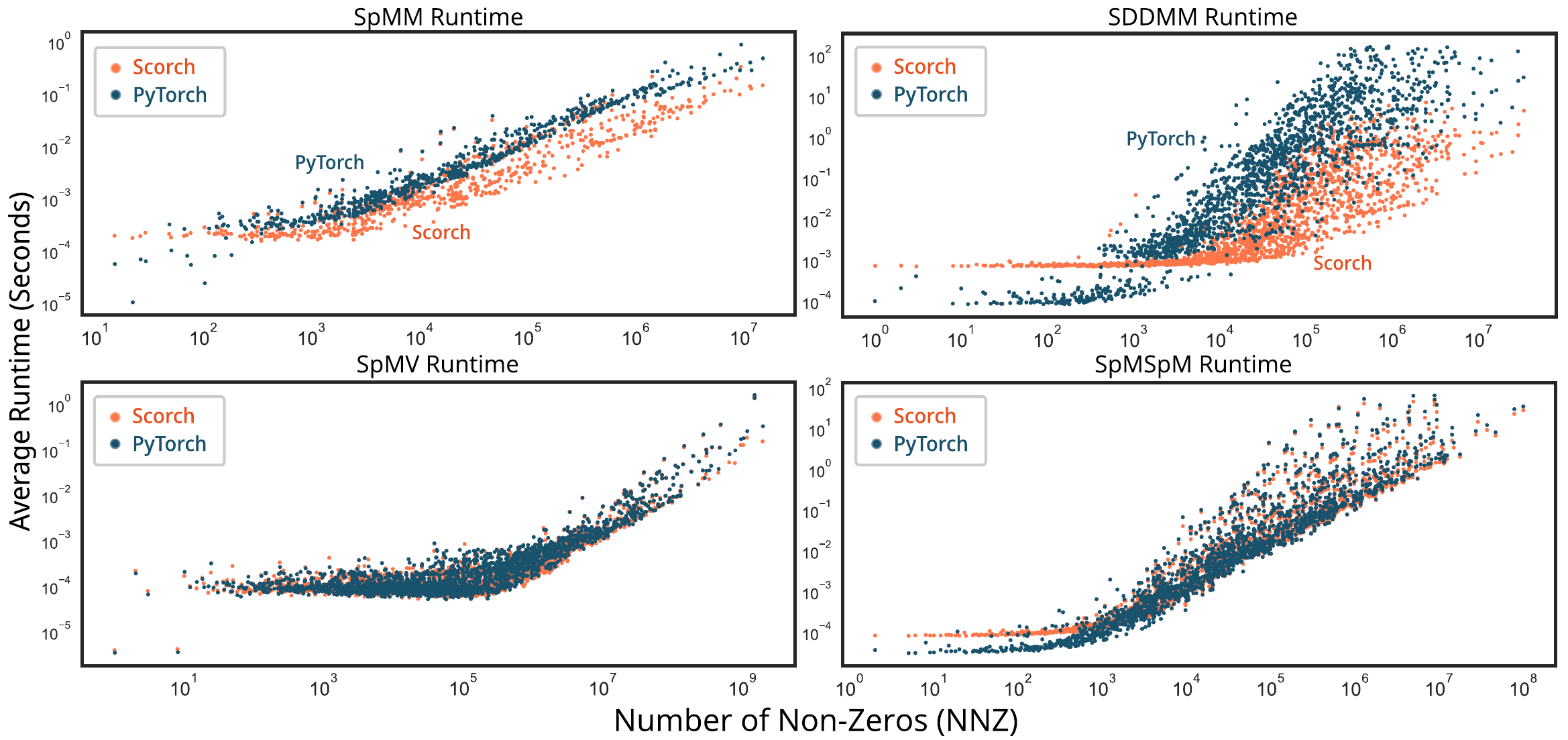}
  \end{center}
  \caption{Performance on key sparse operations}
  \label{fig:all-perf}
\end{figure}

\Cref{fig:all-perf} shows the absolute runtime of \sys and \texttt{torch.sparse} for the four kernels across a range of matrix sizes and sparsity levels. For SpMV and SpMSpM, \sys achieves performance similar to PyTorch as these operations have limited opportunities for optimization. However, for SpMM, \sys exhibits better scaling and outperforms PyTorch for all but the smallest problems ($<10^2$ nonzeros). \sys's ability to generate efficient kernels becomes more advantageous as the problem size grows.
More significant speedups are observed for SDDMM, where \sys is orders of magnitude faster than PyTorch on bigger problems ($> 10^3$ nonzeros). PyTorch cannot fuse the SDDMM operation and must perform a dense matrix multiplication followed by an element-wise multiplication with a sparse matrix. In contrast, \sys is able to generate a single fused kernel that is asymptotically faster.

These results demonstrate the performance benefits of \sys's sparse code generation and optimizations, especially for operations like SDDMM where there are more opportunities for optimization. \sys relieves users of the burden of handwriting and hand-optimizing sparse kernels while offering competitive performance. \Cref{sec:experimental-details} contains additional details on the experimental setup.

\section{Limitations and Conclusion}

\label{limitations}
\textbf{Limitations.} While \sys supports a wide range of sparse tensor operations, it does not yet provide complete coverage of all possible operations. In particular, \sys does not yet support automatic differentiation with sparse tensors, which is a problem outside the scope of this paper but a direction for future work to enable end-to-end training of sparse models.
The current implementation of the compiler stack supports lowering to CPU-optimized C++ code. Additional performance may be achieved by lowering to Triton or CUDA code, and GPU code generation is left as future work.

\textbf{Conclusion.}
\sys is a PyTorch-based framework that accelerates sparse deep learning by providing efficient sparse computation capabilities and seamless integration with PyTorch. \sys enables users to leverage the benefits of sparsity with minimal modifications to their existing code. Experimental evaluation demonstrates \sys's effectiveness in accelerating sparse ML workloads across various domains, achieving significant speedups over PyTorch Sparse.
With its generality and seamless PyTorch integration, \sys makes using sparsity in deep learning more accessible and paves the way for more exploration of efficient sparse models.

\begin{ack}
This work is supported in part by PRISM, one of seven centers in JUMP 2.0, a Semiconductor Research Corporation (SRC) program sponsored by DARPA, by the Swedish Research Council (Grant No. 2018-04329), and by Digital Futures. Alexander J. Root is supported by an NSF Graduate Research Fellowship.
We thank James Dong, Christophe Gyurgyik, Scott Kovach, Rubens Lacouture, Shiv Sundram, and Rohan Yadav for valuable discussion and feedback on early drafts of the paper.
\end{ack}

\bibliography{main}
\bibliographystyle{plainnat}

\newpage
\appendix

\section{Compilation}

\sys is designed as a layer on top of PyTorch that adds sparse functionality to existing operations. \sys intercepts calls to PyTorch operations and dispatches them to either the traditional PyTorch infrastructure or to the Scorch compiler.

\sys dispatches dense computations to the existing PyTorch computing infrastructure. This includes both tensor operations that operate on fully dense tensors and the fully dense sub-computations of blocked-sparse operations.  By reusing the existing dense functionality, we make it easier for users to transition to \sys, as dense operations continue to execute in the same execution environment and with the same performance as before. Moreover, we reduce the complexity of the \sys compiler, as it does not need to be tuned to match existing dense computing libraries on completely dense operations.

For kernels with at least one sparse tensor, \sys first performs format inference to decide the format of the output tensor. Then, \sys generates a fused loop-level intermediate representation (IR) along the lines of TACO's Concrete Index Notation IR~\cite{kjolstad2019workspaces}. The code generation algorithm for the loop-level IR also optimizes loop ordering, as described in \Cref{sec:loop-ordering}, inserting intermediate temporary tensors as necessary. Unlike in dense tensor computations, the loop order of a sparse tensor computation affects its asymptotic complexity~\cite{ahrens_2022}. That is, changing a matrix multiplication from an inner product algorithm to a linear combination of row algorithm can lead to an arbitrarily large speedup. Loop ordering optimization is therefore critical and one of the contributions in \sys is an algorithm that can quickly find a loop order that avoids a bad asymptotic complexity.

Next, \sys tiles any dense loops inside sparse computations to improve cache utilization. Such loops appear because a sparse tensor computation often also contains dense tensors (e.g., SpMM and SDDMM) and because a sparse tensor may have some dimensions that are stored in a dense array (e.g., the first dimension of a CSR matrix is dense). The tiling algorithm for dense loops in \sys (\Cref{sec:tiling}) is designed to avoid introducing costly random searches in sparse data structures. For example, \sys does not tile sparse loops, although that can sometimes lead to performance improvements, because that would introduce costly searches in sparse data structures. It is difficult for a compiler to determine whether the cache benefits outweighs the searching cost. We therefore designed \sys to not tile sparse loops so that the compiler produces predictable good performance.

Finally, the tiled loops are lowered to C++ code, JIT compiled and linked via PyTorch's custom C++ extensions loader. A high-level description of this process is provided in \Cref{alg:compile}.

\begin{algorithm}[h]
  \caption{\sys Kernel Compilation}
  \label{alg:compile}
  \begin{algorithmic}[1]
    \State \textbf{Input:} Tensor expression ${E}$ with input tensors $\mathcal{T}$ and untyped output tensor $T_{out}$.
    \If{all $\mathcal{T}$ are dense}
    \State \Return \texttt{torch.compile}(${E}$) \Comment{Dynamic dispatch for dense kernels.}
    \EndIf
    \State $T_{out}$.format $\leftarrow$ \textsc{InferFormat}(${E}$, $\mathcal{T}$) \Comment{\Cref{app:format-inference}}
    \State $\mathcal{L} \leftarrow$ \textsc{CompileToLoops}(${E}$, $\mathcal{T}$, $T_{out}$) \Comment{\Cref{alg:looporder}}
    \State $\mathcal{L}_{\mathrm{tiled}} \leftarrow $ \textsc{TileOperations}(${E}$, $\mathcal{T}$, $\mathcal{L}$) \Comment{\Cref{alg:tiling}}
    \State \Return \textsc{LowerToC}($\mathcal{L}_{\mathrm{tiled}}$)
  \end{algorithmic}
\end{algorithm}

The current implementation of \sys compiles a single user-provided expression at a time and does not fuse across expressions. We leave extending our compiler infrastructure to fuse sparse operations across expressions. Such work could, for example, use a tracing facility such as the one introduced in PyTorch 2.0~\cite{ansel_2024} to capture the computational graph that serves as the basis for global optimization.

\newpage
\section{Implementation Details}

Scorch is implemented as a Python library, with its compiler written in Python as well. The \sys compiler generates C++ code and pybind bindings, which are then compiled and loaded using PyTorch's custom C++ extension machinery.

\subsection{Sparse Tensor Representation}

Scorch represents sparse tensors using a unified abstraction that can represent different underlying physical sparse storage formats like COO, CSR, and DCSR. This allows specifying computations independently of storage details, leaving the task of generating low-level code that interacts with specific data structures to the compiler.

The \sys implementation has two primary classes that manage sparse tensors. A tensor class contains general information about the tensor that is independent of specific storage, such as its name, shape, and component type. The tensor class also has a reference to an object of a storage class. A tensor storage contains the physical storage of a tensor, including a storage descriptor (i.e., a format), arrays storing sparse indices that identify the stored (typically non-zero) values of a tensor, such as coordinate lists and CSR row pointers, and an array containing the stored values of the tensor.

\subsection{Sparse Tensor Operations}

Scorch supports arbitrary tensor contractions and element-wise operations on sparse tensors. This includes, but is not limited to:
\begin{itemize}
  \item sparse matrix-vector multiplication,
  \item sparse matrix-matrix multiplication,
  \item tensor contractions,
  \item element-wise sparse addition, subtraction, multiplication, and
  \item fused operations.
\end{itemize}

At the core, \sys can compile generalized \texttt{einsum} contraction operations. The data structures that store each tensor in these operations can be determined separately. The code generator builds on prior work on sparse tensor algebra code generation from the TACO line of work~\cite{taco_2017,kjolstad2019workspaces}. Such code generators generate code that co-iterates over any number of data structures stored in different formats. Thus, they support generating fused code that operates on tensors stored in disparate formats.

\subsection{Sparse Temporary Workspaces}

One key feature in \sys that enables general sparse computations is the support for sparse temporary tensors, often called sparse workspaces~\cite{zhang2024compilation}. Such temporaries are useful in kernels where the loop order causes scattering into the result tensor, but where the result tensor data structure does not support random inserts (e.g., a linear combination of rows matrix multiplication with a CSR result). The sparse temporary tensors in \sys allow computations to scatter and gather results in arbitrary order into a sparse output tensor of any format, while also offering significant memory efficiency by storing temporary intermediate tensors in a compressed format, avoiding the need to materialize intermediate zeros values as in dense tensors. In Scorch, a red-black tree is used to store intermediate non-zero values. We chose red-black trees because they are simple and inherently provide the features we need, namely the ability to insert coordinates-value pairs and the ability to iterate over them in sorted order. Workspaces improve performance by avoiding unnecessary format materialization and compression.

\newpage
\section{Auto-scheduling Algorithms}

\subsection{Tiling}
\label{app:tiling-algo}

The tiling algorithm analyzes the tensor expression and determines which loops to tile based on several key observations, as described in \Cref{sec:tiling}. The pseudocode for the tiling algorithm is provided in \Cref{alg:tiling}.

\begin{algorithm}[H]
  \caption{Tiling Sparse Tensor Operations}
  \label{alg:tiling}
  \begin{algorithmic}[1]
    \State \textbf{Input:} A tensor expression $E$ containing input tensors $\mathcal{T}$, and a loop structure $\mathcal{L}$.
    \State \textbf{Output:} Tiled loop structure $\mathcal{L}_{\mathrm{tiled}}$ for efficient computation.
    \State $\mathcal{W} \leftarrow \emptyset$ \Comment{Initialize the working set of index variables}
    \State $\mathcal{S} \leftarrow \textsc{GetIndexVariables(}E\textsc{)}$ \Comment{Get the set of all index variables in the expression}

    \For{each tensor access $T[\mathcal{I}]$ in the expression $E$}
      \If{$\mathcal{I} \subset \mathcal{S}$} \Comment{Check if indices of $T$ are a strict subset of $\mathcal{S}$}
        \State $\mathcal{W} \leftarrow \mathcal{W} \cup \mathcal{I}$ \Comment{Add indices to the working set}
      \EndIf
    \EndFor

    \ForAll{$i \in \mathcal{W}$} \Comment{Remove sparse index variables}
      \If{there exists a tensor $T \in \mathcal{T}$ where the dimension of $T$ corresponding to $i$ is sparse}
        \State $\mathcal{W} \leftarrow \mathcal{W} \setminus \{i\}$
      \EndIf
    \EndFor

    \ForAll{$i \in \mathcal{W}$} \Comment{Remove index variables that are parents of sparse dimensions}
      \If{there exists a sparse index variable $j$ such that $i$ is a parent of $j$ in the loop nest $\mathcal{L}$}
        \State $\mathcal{W} \leftarrow \mathcal{W} \setminus \{i\}$
      \EndIf
    \EndFor

    \ForAll{$i \in \mathcal{W}$} \Comment{Tiling}
      \State $(i_{\text{outer}}, i_{\text{inner}}) \leftarrow \textsc{TileIndex}(i)$ \Comment{Split loop $i$ into outer and inner loops}
      \State $\mathcal{L} \leftarrow \textsc{ReorderLoops}(\mathcal{L}, i_{\text{outer}}, 0)$ \Comment{Reorder $i_{\text{outer}}$ to outermost in the loop nest}
    \EndFor

    \State \Return $\mathcal{L}_{\mathrm{tiled}} \gets \mathcal{L}$
  \end{algorithmic}
\end{algorithm}

\section{Tensor Formats and Format Inference}
\label{app:format-inference}

Scorch uses the sparse tensor format description language proposed by \citet{taco_2017} and \citet{format_2018}. When users create a PyTorch tensor, it is dense by default as it is in vanilla PyTorch. Scorch extends the PyTorch tensor constructor to allow the user to specify that a tensor should be sparse. Sparse tensors are stored in the coordinate format by default, but also lets users specify a specific tensor formats. The Scorch tensor format description language is flexible and support many different types of data structures, such as compressed vectors, compressed matrices, compressed tensors, blocked-sparse matrices and tensors, dense tensors, and coordinate matrices and tensors.

Although users must specify whether a tensor defined using a constructor is sparse, Scorch automatically infers the (dense or sparse) data structures of tensors that result from some operations. So if the user multiplies two sparse matrices (SpGEMM),  Scorch will infer that the result should be stored in a sparse matrix, as well as the specific sparse data structure. And if the user multiplies a sparse matrix with a dense matrix (SpMM), Scorch automatically infers that the result should be stored in a dense array.

\subsection{Tensor Formats}

A tensor format describes the data structures that store the non-zero elements of a tensor. Following \citet{taco_2017}, we allow a separate description of the data structure of each dimension of a tensor. So a compressed sparse row (CSR) matrix stores the set of rows in a dense data structure and each row in a compressed data structure, while a doubly-compressed sparse row (DCSR) matrix stores both dimensions in a compressed data structure. We support three types of per-dimension data structures that can be composed any way to store a tensor of any dimensionality:
\begin{itemize}
  \item \textit{dense}: All elements, including zeros, are explicitly stored.
  \item \textit{compressed}: Non-zeros are stored using compressed index arrays storing their coordinates.
  \item \textit{coordinate}: Non-zeros are stored as a list of coordinate-value pairs.
\end{itemize}

The full format of a tensor is determined by both the ordering of dimensions and the data structure for each dimension. For example, a matrix (2D tensor) can be stored in many ways: a row-major dense matrix (dense rows, dense columns), a row-major compressed sparse row (CSR) matrix (dense rows, compressed columns), a column-major compressed sparse column (CSC) matrix (dense columns, compressed rows), a doubly compressed sparse row (DCSR) matrix (compressed rows, compressed columns), and a row-major coordinate list (COO) matrix (coordinate rows, coordinate columns).

Our way to specify coordinate tensors simplifies their specification in prior work by \citet{format_2018}. They introduced a singleton data structure that needed to be composed with a duplicate version of a compressed data structure, so a coordinate matrix became (unordered compressed rows, singleton columns). Scorch, on the other hand, represents coordinate tensors more uniformly using a coordinate data structure, so a coordinate matrix becomes (coordinate rows, coordinate columns).

\subsection{Format Inference Algorithm}
Given a tensor computation expressed as an einsum operation or as a general tensor algebra expression, we provide an algorithm to infer the format of the output tensor based on the formats of the input tensors. Our inference algorithm determines the storage format for each level of the output tensor independently and thus naturally extends to tensors of any dimensionality.

The inference algorithm traverses the tensor expression and predicts the format of the result by separately predicting the data structure of each of its dimensions. This is done using basic algebraic reasoning: the element-wise multiplication of a sparse vector by any other vector type should be sparse, since the result will be at least as sparse as either operand. And the element-wise addition of a dense vector with any other vector type should be dense, since the result will be at least as dense as either operand. Furthermore, when two sparse vectors are added, Scorch makes the result sparse. Although the result is the union of the operands, and thus denser than  either operand, we elect to make Scorch conservative. The reason for such conservativeness is that instantiating a dense tensor when the result is very sparse can asymptotically increase memory and compute usage, while instantiating a sparse tensor when the result is sparse only increases memory and compute usage by a constant factor.  Finally, when two dense vectors are multiplied, Scorch makes the result dense. We provide the inference algorithm in \Cref{alg:infer}, where \textit{compressed} and \textit{coordinate} formats are both considered \textsc{Sparse}.

\begin{algorithm}
  \caption{Tensor Format Inference}
  \label{alg:infer}
  \begin{algorithmic}[1]
    \Procedure{Infer}{$E$, $lvl$}
    \State \textbf{match} E
    \State | $A + B$ $\rightarrow$ \textbf{match} \textsc{Infer}($A$, $lvl$), \textsc{Infer}($B$, $lvl$) \textbf{with}
    \State \quad \quad \quad \quad \; \;| \textsc{Dense}, \_ $\rightarrow$ \Return \textsc{Dense}
    \State \quad \quad \quad \quad \; \;| \_, \textsc{Dense} $\rightarrow$ \Return \textsc{Dense}
    \State \quad \quad \quad \quad \; \;| \_, \_ $\rightarrow$ \Return \textsc{Sparse}
    \State | $A * B$ $\rightarrow$\; \textbf{match} \textsc{Infer}($A$, $lvl$), \textsc{Infer}($B$, $lvl$) \textbf{with}
    \State \quad \quad \quad \quad \; \;| \textsc{Sparse}, \_ $\rightarrow$ \Return \textsc{Sparse}
    \State \quad \quad \quad \quad \; \;| \_, \textsc{Sparse} $\rightarrow$ \Return \textsc{Sparse}
    \State \quad \quad \quad \quad \; \;| \_, \_ $\rightarrow$ \Return \textsc{Dense}
    \State | $T$ $\rightarrow$ \textsc{TensorFormat}($T$, $lvl$)
    \State \textbf{end}
    \EndProcedure
  \end{algorithmic}
\end{algorithm}

\subsection{Example}
Consider the following tensor computation:
\begin{equation}
  D_{ij} = \sum_k A_{ik} B_{kj} + C_{ij}
\end{equation}

Suppose the input tensors have the following formats:
\begin{itemize}
  \item $A$: CSR (dense $i$, compressed $k$)
  \item $B$: CSR (dense $k$, compressed $j$)
  \item $C$: DCSR (compressed $i$, compressed $j$)
\end{itemize}

To infer the format of the output tensor $D$, we apply the format inference algorithm:

\begin{enumerate}
  \item First, we consider the multiplication sub-expression $T_{ij} = \sum_k A_{ik} B_{kj}$:
  \begin{itemize}
    \item For $T[i]$: dense $A[i]$ contributes through multiplication, so $T[i]$ should be dense.
    \item For $T[j]$: compressed $B[j]$ contributes through multiplication, so $T[j]$ should be compressed.
  \end{itemize}

  \item Next, we consider the addition of $T$ and $C$ to form the final output $D$:
  \begin{itemize}
    \item For $D[i]$: dense $T[i]$ contributes through addition, while compressed $C[i]$ contributes through addition. Since $T[i]$ is dense, $D[i]$ should be dense.
    \item For $D[j]$: compressed $T[j]$ contributes through addition, while compressed $C[j]$ contributes through addition. Therefore, $D[j]$ should be compressed.
  \end{itemize}
  Therefore, the final output tensor $D$ should be stored in CSR format.
\end{enumerate}

This example demonstrates how the format inference algorithm handles expressions with a combination of multiplications and additions by recursively applying the inference rules and combining the results.

\subsection{Discussion}

The tensor format inference algorithm is implemented in the Scorch framework as part of the compiler pipeline. When a tensor computation is encountered, Scorch analyzes the formats of the input tensors and applies the format inference algorithm to determine an appropriate format for the output tensor(s).

The inferred formats are then used to guide the code generation process, ensuring that the appropriate sparse or dense kernels are generated. This automatic format inference capability lets Scorch optimize tensor computations based on the data structures of the input data, resulting in improved memory efficiency and computational performance.

By integrating format inference into the compiler workflow, Scorch enables the seamless mixing of sparse and dense tensors in deep learning models, without requiring manual specification of the result of each and every compute operation. This simplifies the development process and lets users focus on the high-level logic of their models while leveraging the benefits of sparse computations where appropriate.

One limitation of the current algorithm is that it does not consider potential trade-offs between storage efficiency and computational efficiency. In some cases, it may be beneficial to use a denser format for the output tensor to enable more efficient computation, even if it requires more storage. Extending the format inference algorithm to take into account these trade-offs is an interesting direction for future work.

Overall, format inference is a key component of \sys that enables efficient and transparent sparse tensor computation. By automatically inferring the optimal sparse format, \sys simplifies the user experience and ensures high performance across a wide range of sparse tensor operations and sparsity patterns.

\newpage
\section{Experimental Details}
\label{sec:experimental-details}

All experiments were performed on an Apple M1 Ultra CPU (3.2 GHz, 20 cores) with 64 GB of memory.

\subsection{Standard Kernels Benchmark Details}
\label{sec:standard-kernels-details}

We run each benchmark 10 times and report the average runtimes in \Cref{fig:all-perf}. For SpMV, we perform a matrix-vector multiplication of each SuiteSparse matrix with a randomly generated dense vector. For SpMM, we perform the matrix multiplication of each SuiteSparse matrix with a randomly generated dense matrix. For SpMSpM, we truncate any non-square SuiteSparse matrices to be square and matrix multiply them with their transpose. For SDDMM, we perform the element-wise multiplication of each SuiteSparse matrix with the matrix multiplication of two randomly generated dense matrices.

\paragraph{Matrix Formats and Characteristics.} For the SpMV, SpMM, and SDDMM benchmarks, we use sparse matrices in the Compressed Sparse Row (CSR) format, which is widely supported across sparse libraries. However, for SpMSpM, PyTorch does not support matrix multiplication between two CSR matrices on non-Intel machines due to the absence of the Intel Math Kernel Library (MKL). The specific error message is: ``addmm: computation on CPU is not implemented for SparseCsr + SparseCsr @ SparseCsr without MKL.''

To work around this limitation and still evaluate SpMSpM performance, we use the COO (Coordinate) format for SpMSpM. Additionally, since not all matrices in the SuiteSparse collection are square, we truncate non-square matrices to be square and multiply them with their transpose to obtain a valid SpMSpM operation.

\paragraph{PyTorch Performance on Small SpMM Problems.} We observe that PyTorch Sparse outperforms \sys on SpMM for small problems with fewer than $10^2$ nonzeros. This is because PyTorch's SpMM implementation is optimized for small matrices. Their gather-scatter approach is more efficient for small inputs but does not scale well to larger matrices. \sys's generated kernels have some overhead that dominates for small problems, but they scale much better to larger matrices, resulting in the speedups seen in \Cref{fig:all-perf}.

\subsection{Graph Neural Networks}

We evaluate the inference performance of Graph Convolutional Networks (GCNs)~\cite{kipf2016semi} on four node classification datasets: Cora~\cite{cora_2000}, Citeseer~\cite{citeseer_1998}, PubMed~\cite{pubmed_2008}, and OGBN-arXiv~\cite{ogbn_2020}. The model architecture and hyperparameters are listed in \Cref{tab:gcn}. We use the Adam optimizer with a learning rate of 0.01 and weight decay of 5e-4 to train the model for 200 epochs with a batch size of 1 (full-batch). Dropout with a rate of 0.5 is applied to the hidden layer during training.

For each dataset, we train the models on the training set and evaluate the inference time and accuracy on the test set using an Apple M1 Ultra CPU with 64 GB of memory. The experiments are run in PyTorch 2.2.1, PyTorch Geometric 2.5.0, and DGL 2.0.0, and Scorch 0.1.0.
We run each inference experiment 50 times with 5 warm-up runs and report the average speedups of various frameworks relative to PyTorch, as well as the absolute inference times, in \Cref{fig:gcn-perf}.

To ensure a fair comparison, we use the same GCN architecture across all frameworks. The PyTorch and Scorch implementations use a custom GCN layer, while PyG and DGL use their built-in GCN layers. The same model is trained with PyG and the train weights are loaded into the PyTorch, Scorch, PyG, and DGL models after adjusting for any differences in the parameter shapes and orderings.

\begin{table}[h]
\centering
\begin{tabular}{ll}
\toprule
\textbf{Hyperparameter} & \textbf{Value} \\
\midrule
Hidden channels & 128 \\
Activation function & ReLU \\
Dropout rate & 0.5 \\
\cmidrule{1-2}
Optimizer & Adam \\
Learning rate & 0.01 \\
Weight decay & 5e-4 \\
Batch size & 1 (full-batch) \\
Training epochs & 200 \\
\bottomrule
\end{tabular}
\caption{GCN model architecture and hyperparameters for node classification.}
\label{tab:gcn}
\end{table}

\subsection{Sparse Autoencoders}

We evaluate the performance of sparse autoencoders on four datasets: MNIST \citep{lecun1998mnist}, CIFAR-10 \citep{krizhevsky2009learning}, CIFAR-100 \citep{krizhevsky2009learning}, and CelebA \citep{liu2015deep}. The datasets are preprocessed as follows:

\begin{itemize}
    \item MNIST: The images are converted to tensors and flattened into a 1D vector of size 784 (28$\times$28).
    \item CIFAR-10 and CIFAR-100: The images are converted to grayscale, then to tensors, and flattened into a 1D vector of size 1024 (32$\times$32).
    \item CelebA: The images are converted to grayscale, resized to 64$\times$64, then to tensors, and flattened into a 1D vector of size 4096 (64$\times$64).
\end{itemize}

The sparse autoencoder architecture and hyperparameters are summarized in \Cref{tab:sparse_autoencoder}. The encoder is a sparse linear layer followed by ReLU activation, and the decoder is a dense linear layer followed by sigmoid activation.

\begin{table}[ht]
\centering
\begin{tabular}{ll}
\toprule
\textbf{Component} & \textbf{Details} \\
\midrule
Encoder & Sparse Linear (input size $\rightarrow$ 256) \\
Encoder Activation & ReLU \\
Decoder & Dense Linear (256 $\rightarrow$ input size) \\
Decoder Activation & Sigmoid \\
\cmidrule{1-2}
Loss Function & Mean Squared Error (MSE) \\
Optimizer & Adam \\
Learning Rate & 0.01 \\
Batch Size & 64 \\
Training Epochs & 10 \\
\bottomrule
\end{tabular}
\caption{Sparse autoencoder architecture and hyperparameters.}
\label{tab:sparse_autoencoder}
\end{table}

The models are trained using the mean squared error (MSE) loss and the Adam optimizer \citep{kingma2014adam} with a learning rate of 0.01. We use a batch size of 64 and train the models for 10 epochs. The experiments are conducted on an Apple M1 Ultra CPU with 64GB of memory.

During inference, we measure the average reconstruction loss and the wall-clock time on the test set. The reconstruction loss is computed using the MSE loss between the input images and the reconstructed images. The wall-clock time includes the time taken to convert the input data to a sparse CSR tensor, forward pass through the model, and compute the loss.

The experiments are implemented in PyTorch 2.2.1 and Scorch 0.1.0. To ensure a fair comparison, we use the same model architecture, hyperparameters, and random seed across both frameworks.
We run each inference experiment 50 times with 5 warm-up runs and report the average speedups of Scorch relative to PyTorch, as well as the absolute inference times, in \Cref{fig:sparse_autoencoder_results}.

\subsection{Sparse Transformers}

Sparse transformers are variants of the Transformer architecture \cite{vaswani2017attention} that leverages sparse attention mechanisms to improve computational efficiency and scalability. In this evaluation, we implement BigBird \cite{zaheer2020bigbird}, a specific type of sparse transformer that employs a combination of global, sliding window, and random sparse attention patterns.

BigBird is designed to handle long sequences while maintaining a manageable computational complexity. It achieves this by using a sparse attention mechanism that attends to a subset of tokens in the sequence, rather than attending to all tokens as in the standard transformer. The sparse attention pattern in BigBird consists of three components:

\begin{enumerate}
  \item \textbf{Global attention:} Attend to all tokens in fixed-size blocks at regular intervals.
  \item \textbf{Sliding window attention:} Attend to neighboring tokens within a fixed-size window that slides over the sequence.
  \item \textbf{Random attention:} Attend to a fixed number of randomly selected tokens in the sequence.
\end{enumerate}

By combining these attention patterns, sparse transformers can capture both local and global dependencies while significantly reducing the computational cost compared to the standard Transformer.

Sparse transformers have been successfully applied to various natural language processing tasks, such as text classification, question answering, and language modeling, where the input sequences can be very long. It has also shown promising results in other domains, such as genomics and time series analysis, where the ability to handle long sequences is crucial.

We evaluate the inference performance of the BigBird model \citep{zaheer2020bigbird} on three text classification datasets: AG News \citep{zhang2015character}, IMDB \citep{maas2011learning}, and Yahoo Answers \citep{zhang2015character}. The model architecture and hyperparameters are listed in \Cref{tab:bigbird}. We use the AdamW optimizer \citep{loshchilov2017decoupled} with a learning rate of 0.001 to train the models for 5 epochs with a batch size of 64. The sparse attention is configured to use a block size of 16, 2 global tokens, 2 random blocks, and 2 sliding blocks.

For each dataset, we train the models on the training set and evaluate the inference time on the test set using an Apple M1 Ultra CPU. The experiments are implemented in PyTorch 2.2.1 and Scorch 0.1.0.
We run each inference experiment 10 times with 5 warm-up runs and report the average speedups of Scorch relative to PyTorch, as well as the absolute inference times, in \Cref{fig:bigbird}.

\begin{table}[ht]
\centering
\begin{tabular}{ll}
\toprule
\textbf{Hyperparameter} & \textbf{Value} \\
\midrule
Embedding size & 128 \\
Hidden size & 128 \\
Intermediate size & 256 \\
Number of hidden layers & 2 \\
Number of attention heads & 4 \\
Hidden activation & gelu \\
Attention dropout & 0.1 \\
Hidden dropout & 0.1 \\
Sparse attention block size & 16 \\
Sparse attention global tokens & 2 \\
Sparse attention random blocks & 2 \\
Sparse attention sliding blocks & 2 \\
\cmidrule{1-2}
Optimizer & AdamW \\
Learning rate & 0.001 \\
Batch size & 64 \\
Training epochs & 5 \\
\bottomrule
\end{tabular}
\caption{BigBird model architecture and hyperparameters for text classification.}
\label{tab:bigbird}
\end{table}

\section{Extended Related Work}

\textbf{Sparse Tensor Formats.}
To avoid unnecessary computation and storage of zero values, various compressed sparse tensor formats have been developed. These aim to store only the meaningful non-zero values and indices required to fully represent a sparse tensor. Popular formats that store irregular tensors include compressed sparse row/column storage (CSR/CSC) which compactly stores non-zero values and their row/column indices. Coordinate format (COO) uses tuples of indices and corresponding values. Block sparse formats partition tensors into dense sub-blocks for efficiency. Compressed sparse fiber exploits skewed sparsity patterns by orienting storage along one dimension. These formats are all supported by Scorch.

There are also many tensor formats that are not supported by Scorch. Many of these have been developed to take advantage of additional structure in the tensors, such as ELLPACK, which assumes a bounded number of non-zeros per row, and the diagonal matrix format, which assumes all non-zeros are located on a small number of diagonals. Other formats store additional data to matrices in both row- and column-major order, such as the CSB matrix format~\cite{buluc_2009} and the blocked matrix format used by MegaBlocks~\cite{megablocks2023}. Finally, several tensor formats have been proposed in recent years to further improve the performance of tensor algebra operations at the expense of more complexity in the data structures, such as CSR5~\cite{liu_2015} and HiCOO~\cite{li_2018}. In the current version of Scorch, we chose to focus on the data structures that account for the vast majority of use, to manage engineering complexity. However, we believe it is interesting future work to add more formats into Scorch while supporting any operations on any combination of data structures..

Each format makes different trade-offs between compression, indexing overhead, and read/write efficiency for varied sparsity characteristics. Choosing appropriate sparse storage and operations can significantly reduce computational complexity and memory usage compared to dense defaults. However, current deep learning frameworks lack native support for sparse data structures across all compute operations.

\textbf{Sparse Kernels.}
In addition to compressed storage, many specialized sparse kernels have also been created, including for sparse matrix multiplication (SpMM), sparse matrix-vector multiply (SpMV), sparse convolution, and more. Key optimizations in these kernels include techniques like iteration space tiling, loop reordering to improve locality, vectorization, and load balancing to handle irregular sparsity. Highly optimized sparse libraries like NVIDIA cuSPARSE and Intel MKL provide some essential sparse operations, but composing and scheduling sparse kernels in deep learning frameworks remains a challenging problem.

Despite the potential benefits of sparsity, adoption in deep learning still faces multiple key challenges: the lack of versatile software frameworks to natively exploit sparsity in the data and models; the mismatch between existing dense architectures like GPUs/TPUs and sparse computation; the difficulty of fusing sparse operations with other layers in an end-to-end model; the irregular memory access patterns of sparse kernels impairing performance compared to dense routines; the increased difficulty of parallelization and load balancing due to sparse connectivity; and finding the right abstraction level to balance performance and productivity.

\textbf{Auto-scheduling.}
\sys contributes techniques for automatically determining optimal schedules for sparse kernels. These are related to auto-scheduling approaches in compilers.
The automatic scheduling of computational kernels is challenging due to the exponential search space of possible optimizations like loop transforms and data layouts.
Prior auto-scheduling techniques for dense~\cite{halide_auto_2019,flextensor_2020,ansor_2020} and sparse tensors~\cite{ahrens_2022,kanakagiri_2023} rely on cost modeling, optimization heuristics, and black-box search.
Many auto-schedulers for dense tensor programs exist, such as the Halide auto-scheduler~\cite{halide_auto_2019}, FlexTensor~\cite{flextensor_2020}, and Ansor~\cite{ansor_2020}.
Some recent work has focused on auto-scheduling for tensor algebra.
CIN-P~\cite{ahrens_2022} enumerates schedules based on asymptotic costs to find optimal schedules for general sparse tensor algebra expressions.
SpTTN-Cyclops~\cite{kanakagiri_2023} tunes contraction paths and index orders for SpTTN (contractions of a single large sparse tensor with several dense tensors) kernels using both enumeration and efficient search.
\sys contributes a lightweight heuristic-based auto-scheduler for general sparse kernels that leverages the structure of sparse tensor operations to efficiently explore the search space of different tensor kernels. Moreover, we show in our evaluation that the Scorch heuristic auto-scheduler is able to find good implementations of many operations.

\textbf{Sparsity in Deep Learning.}
Sparsity is inherent in many emerging domains for deep learning, including recommender systems~\cite{DLRM19}, drug discovery~\cite{Zhang2019SFLLNAS,elbadawi_2021}, and web-scale graph analytics~\cite{hamilton_2017}. For example, in recommendation systems, user-item interaction matrices are often sparse, as each user interacts with a small subset of the catalog. Molecular graph representations are similarly sparsely connected. Large-scale graphs such as web and social network graphs also exhibit power law distributions, resulting in highly sparse adjacency matrices.

Despite this sparsity, the default approach in deep learning relies heavily on dense matrix multiplication and convolution operations~\cite{sze2017efficient}. Using dense matrix multiplication algorithms on extremely sparse data wastes both computation and memory.
The massive computational and memory burdens incurred by ignoring sparsity quickly become prohibitive as data continues growing in scale and dimensionality~\cite{hoefler2021sparsity}.

Several works have developed sparse neural network architectures.
Methods like sparse evolutionary training~\cite{mocanu2018scalable} and dynamic sparse reparameterization~\cite{mostafa_2019} induce sparsity during training through pruning or regularization.
The lottery ticket hypothesis~\cite{frankle_2019} shows that networks can be trained from scratch to be sparse.
Structured sparsity techniques induce sparsity in regular patterns, such as block sparsity and channel sparsity, to optimize memory access and computation for specific hardware architectures~\cite{wen2016learning}.
While these methods demonstrate the potential benefits of sparsity in deep learning, they often require custom sparse kernels to be implemented for the specific sparse operations used~\cite{gale2020sparse}.
Scorch enables researchers to more easily explore novel sparse architectures by providing a general framework for efficient sparse computation. Rather than needing to implement custom kernels, users can rely on Scorch's compiler to automatically generate performant code for their particular sparse operations. This reduces the engineering burden and allows research to focus on modeling innovations.

\end{document}